\documentclass[
hf,
]{ceurart}

\usepackage{listings}
\usepackage{graphicx} %
\usepackage{subcaption}
\usepackage{caption}
\usepackage{multirow}
\usepackage{hyperref}
\usepackage{booktabs}
\usepackage[table]{xcolor}
\usepackage{float}
\usepackage{pgfplots}
\usepackage{pgfplotstable}
\pgfplotsset{compat=1.17}
\usepackage{adjustbox,xurl}

\usepackage[most]{tcolorbox}
\usepackage[normalem]{ulem}

\newtcolorbox{simplificationexample}{
  enhanced,
  colback=white,
  colframe=black,
  boxrule=0.5pt,
  arc=1pt,
  left=8pt,
  right=8pt,
  top=8pt,
  bottom=8pt,
  before skip=0pt,
  after skip=0pt
}

\newtcolorbox{phaseoneexample}{
  enhanced,
  colback=white,
  colframe=black,
  boxrule=0.5pt,
  arc=1pt,
  left=8pt,
  right=8pt,
  top=8pt,
  bottom=8pt,
  before skip=0pt,
  after skip=0pt
}

\begin{document}

\copyrightyear{2026}
\copyrightclause{Copyright for this paper by its authors.
  Use permitted under Creative Commons License Attribution 4.0
  International (CC BY 4.0).}

\conference{SIG Knowledge Management Workshop (FG WM) at KI 2026, 2026, Bremen, Germany}

\title{A Human-in-the-Loop Corpus for LLM-Based Simplification of Scientific Summaries}

\author{Kyuri Im} %
\author{Michael Färber}[email=michael.faerber@tu-dresden.de]

\address{ScaDS.AI, Technische Universität Dresden, Dresden, Germany}

\begin{abstract}
Interdisciplinary research is accelerating, yet scientific papers remain difficult to read outside their home fields. We study large language model (LLM)-based simplification for scientific texts and present a human-in-the-loop workflow that turns expert summaries into more accessible versions for non-specialists. Using SciSummNet as the source corpus, we first generate baseline simplifications with GPT-4o-mini. In Phase~1, readers from STEM domains outside computer science identify difficult sentences and phrases and compare the original and GPT-simplified summaries in terms of understanding, naturalness, and simplicity. In Phase~2, computer science experts use this feedback to produce expert-edited reference simplifications. We release the resulting corpus together with human judgments and automatic evaluation results. The Phase~1 judgments show a clear preference for the GPT outputs in understanding and simplicity, while qualitative analysis of the Phase~2 edits illustrates the importance of preserving domain terminology and scientific claim strength. The resource supports the training and benchmarking of simplification systems for cross-disciplinary scientific communication.\\
Dataset: \url{https://github.com/faerber-lab/scientific-text-simplification-corpus}
\end{abstract}


\maketitle

\section{Introduction}

Interdisciplinary research is increasingly important, as many scientific advances emerge at the intersection of different fields \cite{advantagesinterdisciplinaritymodernscience,Cunningham_2021}. At the same time, scientific writing is typically tailored to specialists within a particular discipline rather than to researchers from neighboring fields. Domain-specific terminology, highly compressed arguments, and field-specific writing conventions can therefore substantially limit the accessibility, interpretation, and reuse of scientific findings \cite{TSSurvey}.

Large language models (LLMs) provide a promising basis for simplifying scientific and other specialized texts. Their strong zero- and few-shot capabilities have been demonstrated across a broad range of language tasks \cite{fewshot,zeroshot}, and recent studies have explored their use for scientific and specialized text simplification \cite{engelmann2023textsimplificationscientifictexts,guidroz2025llmbasedtextsimplificationeffect}. However, concerns remain regarding hallucinations, loss of nuance, and inconsistent writing styles. Moreover, most existing benchmarks focus on news articles or Wikipedia passages rather than scientific discourse and cross-disciplinary audiences. Consequently, suitable datasets and evaluation protocols for scientific text simplification remain limited \cite{qiang2025redefiningsimplicitybenchmarkinglarge}.

We address this gap through a human-in-the-loop framework that operationalizes accessibility for STEM readers without expertise in the respective source domain while preserving the scientific meaning of the original text. Using SciSummNet \cite{yasunaga2019scisummnetlargeannotatedcorpus} as source material, we (i) generate initial simplifications using an LLM, (ii) collect sentence- and phrase-level difficulty annotations and assessments of understanding, naturalness, and simplicity from STEM readers outside the respective domain, and (iii) create expert-edited reference simplifications informed by this feedback. The resulting design directly addresses the central trade-off between improving readability and preserving scientific precision, which remains difficult for fully automated simplification systems.

Overall, our contributions are as follows:
\begin{itemize}\setlength\itemsep{0.2em}
    \item A human-in-the-loop methodology for simplifying scientific summaries that centers comprehension by cross-disciplinary STEM readers while maintaining technical accuracy.
    \item A curated corpus\footnote{Our code and data are available at \url{https://github.com/faerber-lab/scientific-text-simplification-corpus}.} derived from SciSummNet with (a) baseline LLM outputs, (b) sentence- and phrase-level difficulty annotations, and (c) expert-edited reference simplifications.
    \item An evaluation showing where LLMs improve surface readability and where expert post-editing is crucial for claim calibration and terminology fidelity.
\end{itemize}

The remainder of the paper is structured as follows. Section~2 reviews related work on scientific text simplification and human-centered evaluation. Section~3 describes the corpus construction. 
Section~4 presents the results of the two study phases, followed by limitations and conclusions in Sections~5 and~6.

\section{Related Work}

\textbf{LLMs for Scientific Text Simplification.~}
LLMs have demonstrated strong zero- and few-shot capabilities across a broad range of language tasks \cite{fewshot,zeroshot}. In a zero-shot setting, the model receives only the task instructions, whereas few-shot prompting additionally provides a small number of input-output demonstrations. Recent studies have applied these capabilities to scientific and specialized text simplification \cite{engelmann2023textsimplificationscientifictexts,guidroz2025llmbasedtextsimplificationeffect}. Their results indicate that LLMs can improve accessibility for readers outside the source domain, but risks remain regarding hallucination, loss of nuance, and changes in claim strength. Beyond these quality concerns, most widely used benchmarks are drawn from news or Wikipedia, offering limited coverage of scientific discourse and cross-disciplinary audiences \cite{qiang2025redefiningsimplicitybenchmarkinglarge}. This work addresses that gap by focusing on scientific summaries and collecting difficulty annotations from cross-disciplinary STEM readers.

\textbf{Human-Centered Evaluation in NLP.~}
Automated metrics such as BLEU and ROUGE \cite{papineni2002bleu,lin2004rouge} and readability formulas like Flesch–Kincaid \cite{flesch2007flesch} capture surface overlap and fluency proxies but correlate imperfectly with actual understanding -- especially for technical material. The simplification literature emphasizes evaluating meaning preservation, grammaticality, and fluency beyond n-gram overlap \cite{sulem-etal-2018-bleu}. Complementing this, user-centered assessments target simplicity and reader utility for specific audiences \cite{siddharthan2006syntactic,vajjala2014readability}, as well as discourse coherence and informational adequacy \cite{xu2015problems,kauchak2013improving}. Empirical evidence suggests that LLM-generated simplifications can aid non-specialists, but expert oversight is often required to maintain precision and appropriate calibration \cite{guidroz2025llmbasedtextsimplificationeffect}.

\textbf{Positioning.~}
Our study operationalizes these insights by combining: (i) baseline LLM simplifications; (ii) structured feedback from STEM readers outside computer science on understanding, naturalness, and simplicity; and (iii) expert post-editing to produce reference simplifications. This design directly targets the known trade-off between readability and domain fidelity, and provides a corpus and evaluation setting aligned with real cross-disciplinary use.

\section{Methodology}

Our methodology combines LLM-based simplification with targeted human feedback. It consists of corpus selection, initial simplification, a two-stage user study, and automatic and qualitative evaluation of the resulting simplifications.

\subsection{Dataset}
We adopt SciSummNet \cite{yasunaga2019scisummnetlargeannotatedcorpus}, a corpus of 1,000 highly cited ACL papers with abstracts, citation contexts, and 150-word expert-written summaries. Although these summaries capture the central contributions of the respective papers, they often retain dense terminology and complex syntactic structures. Prior studies suggest that summarization alone does not guarantee readability, while unrestricted simplification can cause verbosity or information loss \cite{10.3389/frai.2024.1375419}. We therefore use the SciSummNet summaries as source texts and simplify them for readers outside computer science while aiming to preserve their scientific content.

\subsection{Baseline Simplification} 
We generate initial simplified summaries using GPT-4o-mini in a zero-shot setting, i.e., without providing input-output demonstrations. The system prompt asks the model to simplify the text for readers without expertise in the scientific field and for the general public, identify complex words or key phrases, and preserve the paragraph format. The user prompt provides the paper title and source summary and asks the model to retain the title while simplifying the summary. The exact prompt is included in the released repository. These LLM outputs serve as the baseline for subsequent human evaluation and refinement.

\subsection{Phase 1: Cross-Disciplinary Reader Evaluation}

\begin{figure}[tb]
    \centering
    \includegraphics[width=0.8\linewidth]{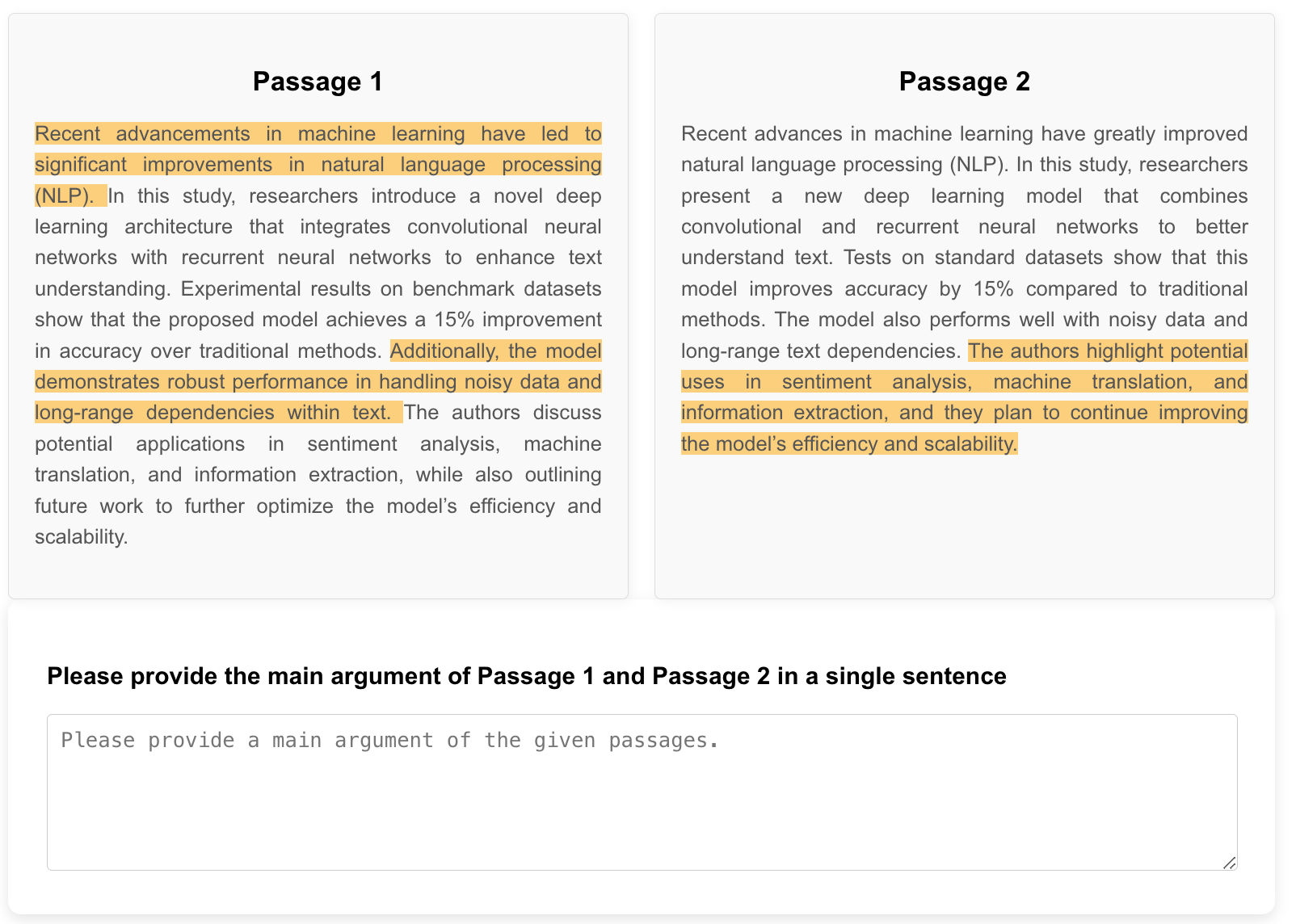}
    \caption{Interface for selecting difficult sentences in Phase~1.}
    \label{fig:survey-phase1}
\end{figure}

\begin{figure}[tb]
    \centering
    \includegraphics[width=0.75\linewidth]{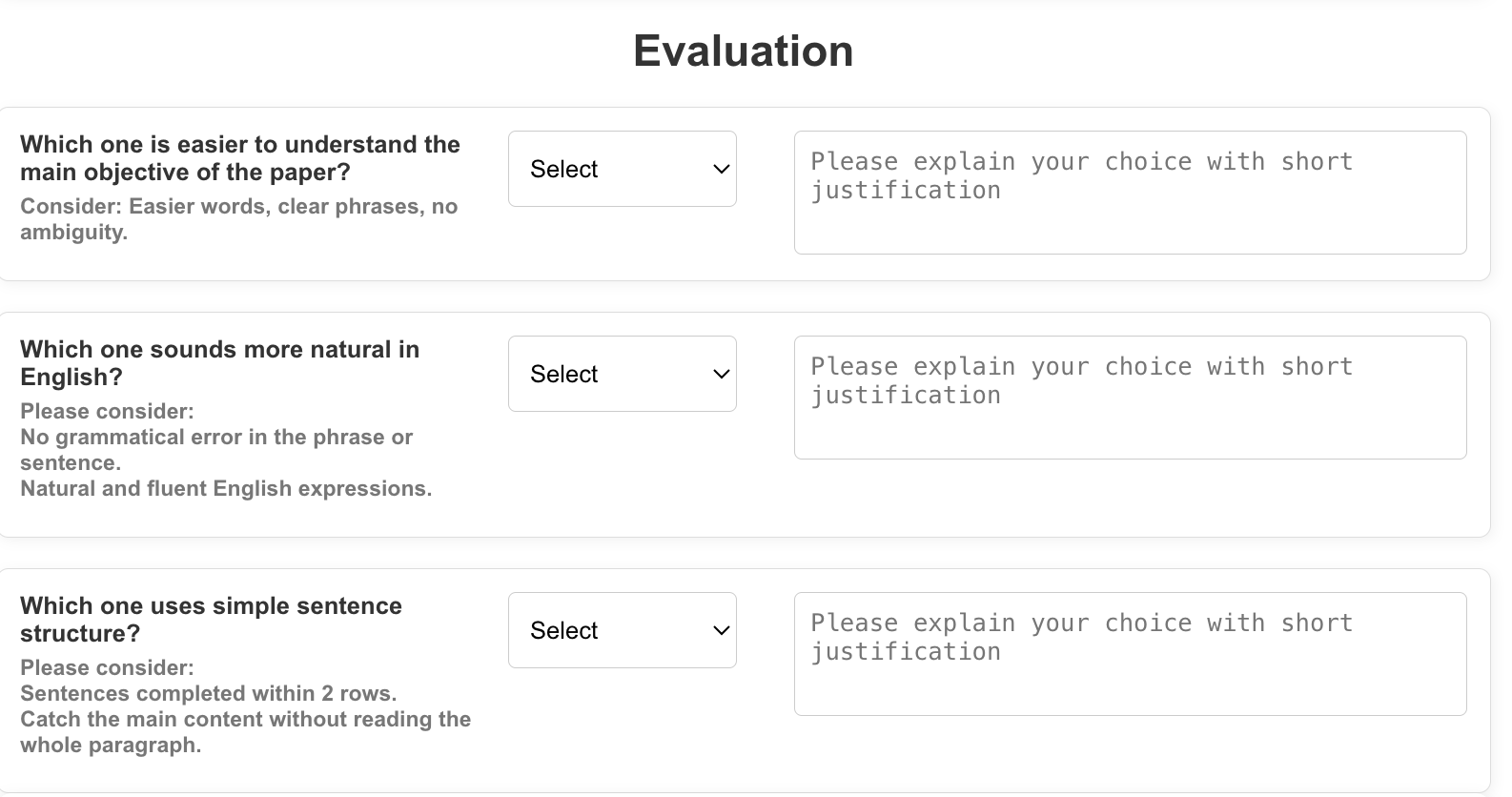}
    \caption{Interface for comparing summaries by understanding, naturalness, and simplicity in Phase~1.}
    \label{fig:eval-question}
\end{figure}

\begin{figure}[tb]
    \centering
    \includegraphics[width=0.7\linewidth]{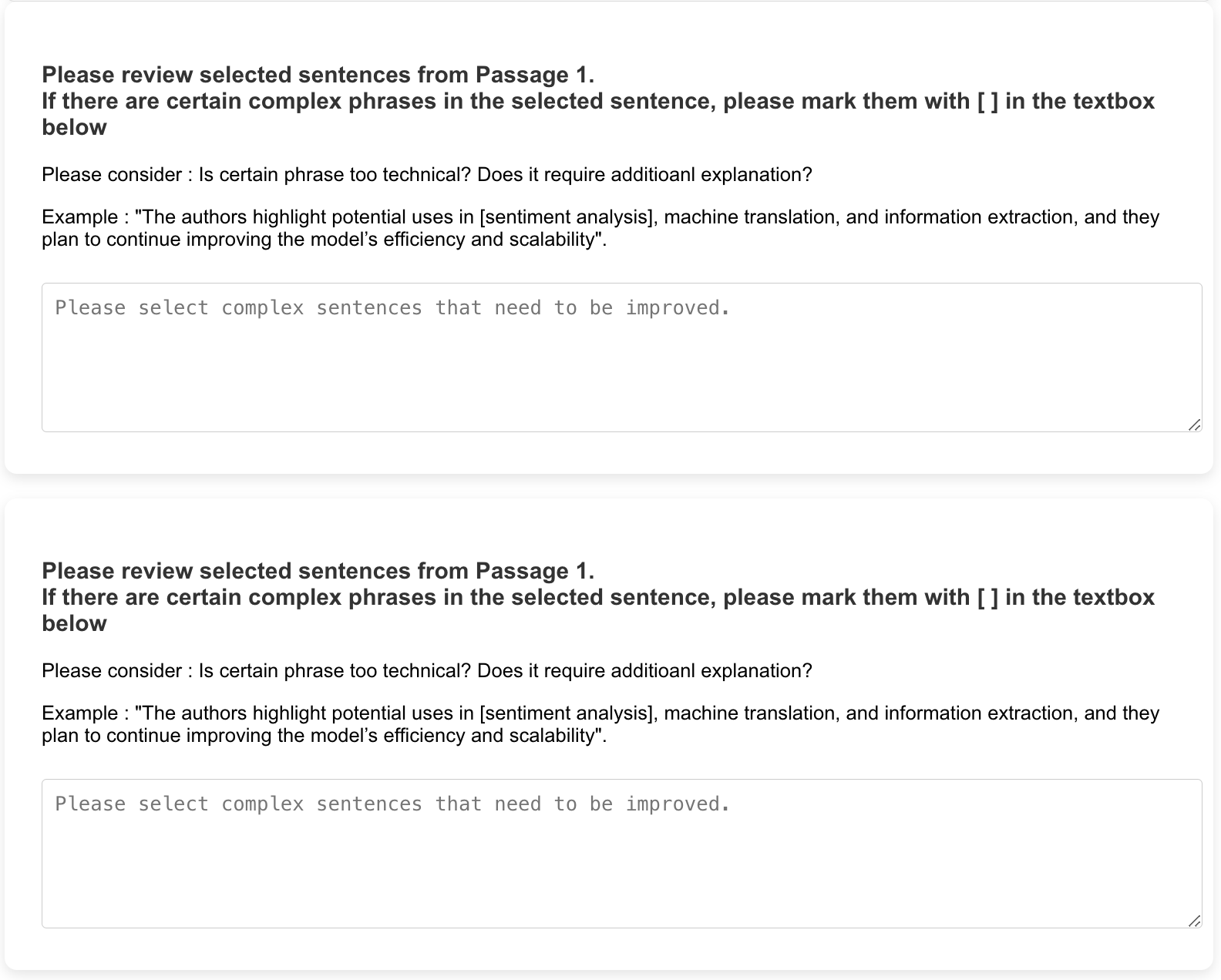}
    \caption{Interface for highlighting difficult words and phrases in Phase~1.}
    \label{fig:annotate-task}
\end{figure}

\textbf{Participants.}
Participants were recruited from STEM disciplines outside computer science and were fluent in English.
\newline
\textbf{Design.}
For each evaluation, a participant was presented with an original SciSummNet summary and its GPT-simplified version in randomized order. Participants (i) selected sentences they found difficult to understand, (ii) made a three-way comparative judgment for \emph{Understanding}, \emph{Naturalness}, and \emph{Simplicity} by preferring the original version, preferring the GPT-simplified version, or indicating no difference, and (iii) highlighted specific words or phrases they considered complex. These are the labels used in the interface; elsewhere, understanding and naturalness are occasionally discussed using the closely related terms coherence and fluency, respectively. Across the completed evaluations, this resulted in 92 comparative judgments per dimension. The annotation interface and task flow are illustrated in Figures~\ref{fig:survey-phase1}--\ref{fig:annotate-task}.

\subsection{Phase 2: Expert-Edited Simplifications}

\begin{figure*}[tb]
    \centering
    \includegraphics[width=0.8\textwidth]{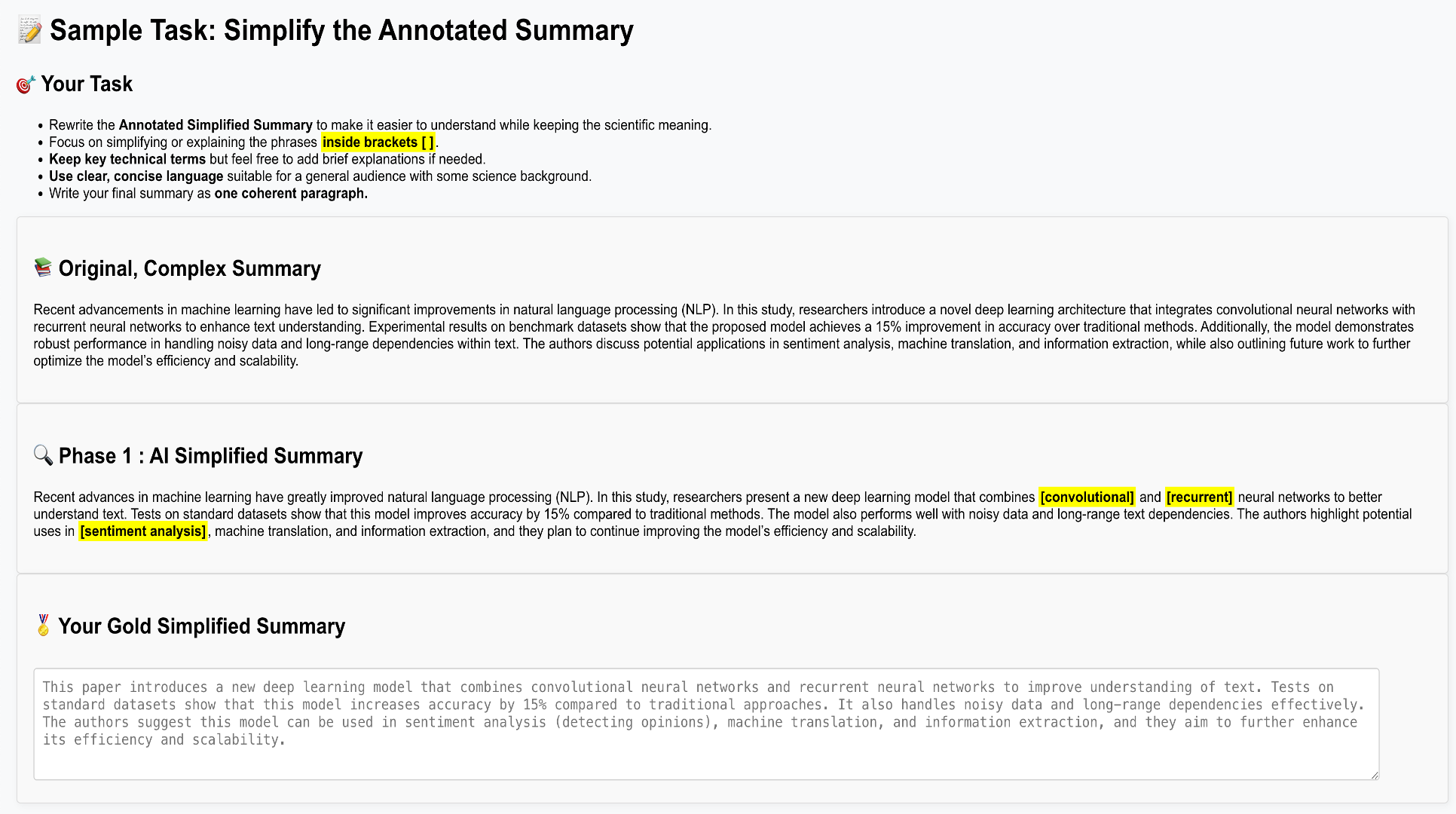}
    \caption{Interface for creating expert-edited reference simplifications in Phase~2.}
    \label{fig:survey-phase2}
\end{figure*}

\textbf{Participants and materials.}
Editors were computer science graduate students or researchers with strong English proficiency. The Phase~2 task covered 92 candidate summaries, of which 47 were completed and included as expert-edited references in the current corpus and evaluation.
For each item, experts received (i) the original summary, (ii) the GPT-simplified version, and (iii) user highlights from Phase~1 indicating difficult words or phrases.
\newline
\textbf{Procedure.} Experts created an expert-edited reference simplification following four guided scenarios:
\begin{enumerate}\setlength\itemsep{0.2em}
    \item \textbf{Both} summaries contain difficult spans: refine and merge content for clarity and fidelity.
    \item \textbf{Only Original} flagged: verify and edit GPT simplifications where necessary.
    \item \textbf{Only GPT} flagged: correct over-simplifications or awkward phrasing.
    \item \textbf{No Flags}: polish structure and verify meaning retention.
\end{enumerate}
Our editing guidelines emphasize retaining technical terminology when required for accuracy, preserving the strength and scope of scientific claims, and using shorter, well-structured sentences where possible. The user interface is shown in Figure~\ref{fig:survey-phase2}.

\section{Results}

We evaluate (i) how cross-disciplinary STEM readers perceive LLM-generated simplifications of scientific summaries in Phase~1 and (ii) how expert-edited reference simplifications compare with the LLM outputs according to automatic metrics in Phase~2. We complement the quantitative results with qualitative analyses of where automated simplification is effective and where expert editing remains important.

\subsection{Phase 1: Cross-Disciplinary Reader Judgments}

\begin{figure}[tb]
\centering
\begin{adjustbox}{max width=0.6\linewidth}
\begin{tikzpicture}
\begin{axis}[
    ybar,
    bar width=12pt,
    enlargelimits=0.15,
    legend style={at={(0.5,-0.15)},anchor=north,legend columns=-1},
    ylabel style={yshift=1em},
    ylabel={Number of Judgments},
    symbolic x coords={Understanding, Naturalness, Simplicity},
    xtick=data,
    nodes near coords,
    nodes near coords align={vertical},
    ymin=0,
    ymax=80,
    width=14cm,
    height=8cm,
    tick label style={font=\small},
    legend cell align={left}
]
\addplot+[fill=blue!60] coordinates {(Understanding,73) (Naturalness,36) (Simplicity,70)};
\addplot+[fill=red!60] coordinates {(Understanding,11) (Naturalness,14) (Simplicity,8)};
\addplot+[fill=gray!50] coordinates {(Understanding,8) (Naturalness,42) (Simplicity,14)};
\legend{GPT-Simplified, Original, No Difference}
\end{axis}
\end{tikzpicture}
\end{adjustbox}
\caption{Comparative judgments across the three Phase~1 evaluation dimensions. Each dimension comprises 92 judgments; no inferential significance test was conducted.}
\label{fig:phase1-bar}
\end{figure}
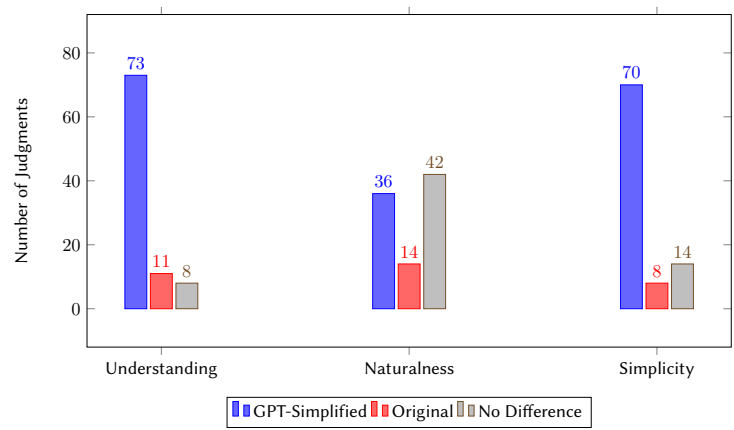

\paragraph{Preferences.}
As shown in Figure~\ref{fig:phase1-bar}, participants favored the LLM-generated summaries for \emph{Understanding} in 73 of 92 judgments and for \emph{Simplicity} in 70 of 92 judgments. This indicates a clear descriptive preference for the simplified versions. Judgments of \emph{Naturalness} were more balanced: 42 indicated no difference, whereas 14 favored the original summary, suggesting that the LLM outputs occasionally contained stylistic inconsistencies or overly informal phrasing.

\paragraph{Difficulty Annotations.}
Sentence- and phrase-level annotations show that participants most frequently flagged multi-noun compounds, unexplained abbreviations, and domain-specific terminology. These annotations informed Phase~2 editing by identifying where simplification should expand compressed terminology, improve discourse flow, or clarify referents.

\paragraph{Qualitative Insights.}
Participant comments and highlights, shown in Figures~\ref{fig:phase1-user-feedback} and~\ref{fig:phase1-user-feedback_2}, reveal two recurring tensions: (1)~\textbf{Precision vs.\ Approachability:} replacing terms such as \emph{latent variables}, \emph{PCFG}, or \emph{NP-hard} with generic paraphrases can improve readability but reduce technical precision; and (2)~\textbf{Style and Naturalness:} the LLM often improves local clarity but occasionally introduces awkward or overly informal formulations, whereas the original summaries generally retain a more conventional academic style.

\begin{figure*}[tb]
\centering

\begin{minipage}{\linewidth}
\begin{phaseoneexample}

\textbf{Naturalness:}
``Passage 1 sounds a little more natural overall.
Its word choices (``arbitrary threshold,'' ``notion of correctness,''
``sentence-level QE system'') match standard academic English and read
smoothly, whereas some substitutions in Passage 2---such as
``idea of correctness is not easy to understand'' and
``chosen limit''---feel slightly informal or awkward. Both passages are
grammatically correct, but Passage 1's vocabulary and phrasing
\textbf{align better with conventional, fluent scholarly style.}''

\vspace{1em}

\textbf{Simplicity:}
Passage 1---its sentences are a little shorter and have
\textbf{fewer parenthetical insertions or stacked phrases}, so each idea
is delivered in a cleaner, more easily digestible structure.

\vspace{1.5em}

\textbf{``Original'':}
``We present a detailed study of confidence estimation for machine
translation. Various methods for determining whether [MT] output is
correct are investigated, for both [whole sentences] and [words]. Since
the [notion of correctness] is not intuitively clear in this context,
different ways of defining it are proposed. We introduce a
[sentence-level QE system] where an \textbf{[arbitrary threshold]} is used
to classify the [MT output] as good or bad. Since the
\textbf{notion of correctness} is not intuitively clear in this context,
different ways of defining it are proposed. We present results on data
from the NIST 2003 Chinese-to-English MT Evaluation. We introduce
\textbf{a sentence level QE system} where an arbitrary threshold is used
to classify the MT output as good or bad. We study sentence and word
level features for translation error prediction.'',

\vspace{1em}

\textbf{``GPT-Simplified'':}
``We present a detailed study of confidence estimation for machine
translation. Various methods for determining whether MT (Machine
Translation) output is correct are investigated, for both whole
sentences and words. Since the
\textcolor{red}{\textbf{[idea of correctness] is not easy to understand}}
in this context, different ways of defining it are proposed. We
introduce a [sentence-level QE (Quality Estimation) system] where a
chosen limit is used to classify the [MT output] as good or bad. We
present results on data from the NIST 2003 Chinese-to-English MT
Evaluation. We introduce a sentence level QE (Quality Estimation) system
where \textcolor{red}{\textbf{a chosen limit}} is used to classify the MT
output as good or bad. We study sentence and word level features for
translation error prediction.''

\end{phaseoneexample}
\end{minipage}

\caption{Example participant feedback on language complexity in scientific summaries.}
\label{fig:phase1-user-feedback}
\end{figure*}

\begin{figure}[t]
\centering

\begin{minipage}{\linewidth}
\small 
\begin{phaseoneexample}

\textbf{Understanding:}
``Passage 1 explains the technical words better.''

\vspace{1.5em}

\textbf{``Original'':}
``This paper defines a generative probabilistic model of parse trees,
which we call PCFG-LA. This model is an extension of PCFG in which
non-terminal symbols are augmented with \textbf{latent variables}.
Fine-grained \textbf{CFG rules} are automatically induced from a parsed
corpus by training a PCFG-LA model using an EM-algorithm. Because exact
parsing with a \textbf{PCFG-LA is NP-hard}, several approximations are
described and empirically compared. In experiments using the Penn WSJ
corpus, our automatically trained model gave a performance of [86.6\%
(F1, sentences $<\!=$ 40 words)], which is comparable to that of an
unlexicalized [PCFG parse]r created using extensive manual feature
selection. We use a markovized grammar to get a better unannotated parse
forest during decoding, but we do not markovize the training data. We
right-binarize the tree bank data to construct grammars with only unary
and binary productions.'',

\vspace{1em}

\textbf{``GPT-Simplified'':}
``This paper describes a method for creating parse trees, which are
structures used in understanding sentences, called PCFG-LA. This method
improves upon a basic version known as PCFG by
\textcolor{red}{\textbf{adding hidden elements}} to the symbols used in
the trees. Detailed rules for constructing these trees are automatically
learned from a collection of example sentences by training the PCFG-LA
model using a technique called the EM-algorithm. Since finding an exact
solution with PCFG-LA is
\textcolor{red}{\textbf{very complex and difficult (NP-hard)}}, the paper
discusses several simplified methods and compares them through
experiments. In tests with a well-known set of sentences (Penn WSJ
corpus), our model, which was trained automatically, performed quite well
with an accuracy of 86.6\% for sentences up to 40 words long, similar to
another model that required a lot of manual adjustment. We use a special
method (markovized grammar) to get better results when interpreting
sentence structures but do not apply this method to the training
examples. We also change the structure of example sentence data to make
it simpler, using only two types of constructions (unary and binary).''

\end{phaseoneexample}
\end{minipage}

\caption{Example participant feedback highlighting clarity issues in scientific summaries.}
\label{fig:phase1-user-feedback_2}
\end{figure}

These observations highlight a central limitation of automated simplification: LLMs can improve surface readability while compromising domain fidelity or stylistic appropriateness. Phase~2 therefore uses expert post-editing to refine the LLM outputs while preserving their scientific meaning.

\subsection{Phase 2: Evaluation of Expert-Edited Simplifications}

\paragraph{Evaluation Setup.}
The evaluation covers the 47 items for which an original SciSummNet summary, an LLM-generated simplification, and a finalized expert-edited simplification are available. In the \emph{GPT-Simplified} condition, each original summary is compared with its LLM-generated version. In the \emph{Expert-Edited} condition, the same original summary is compared with its expert-edited version.

We report four types of automatic measurements. Sentence-level BLEU is calculated for each lowercased and tokenized source-output pair and averaged across the 47 items. BERTScore F1 is calculated using \texttt{distilbert-base-uncased} and likewise averaged across items. SARI is computed without human references by supplying empty references and should therefore be interpreted as a descriptive, reference-free measure of edit operations rather than standard reference-based SARI. Finally, Flesch-Kincaid Grade Level and Reading Ease are computed over the concatenated outputs to characterize corpus-level surface readability. The metric scales and preferred directions are reported in Table~\ref{tab:simplification-eval}.

\begin{table}[tb]
\centering
\begin{adjustbox}{max width=\linewidth}
\rowcolors{2}{white}{blue!5}
\renewcommand{\arraystretch}{1.15}
\begin{tabular}{p{2.8cm}p{5.5cm}cc}
\toprule
\textbf{Category} &
\textbf{Metric and scale} &
\textbf{GPT-Simplified} &
\textbf{Expert-Edited} \\
\midrule
\cellcolor{blue!10}\textbf{Source similarity}
    & Mean sentence BLEU ($0$--$1$, $\uparrow$) & 0.949 & 0.349 \\
\cellcolor{blue!10}
    & Mean BERTScore F1 ($-1$--$1$, $\uparrow$) & 0.9907 & 0.9031 \\
\cellcolor{blue!10}\textbf{Simplification}
    & Reference-free SARI ($0$--$100$, $\uparrow$) & 46.76 & 33.72 \\
\cellcolor{blue!10}\textbf{Readability}
    & FK Grade Level (no fixed bounds, $\downarrow$) & 12.10 & 13.00 \\
\cellcolor{blue!10}
    & FK Reading Ease (no fixed bounds, $\uparrow$) & 48.54 & 40.89 \\
\bottomrule
\end{tabular}
\end{adjustbox}
\caption{Automatic evaluation of 47 aligned source summaries. For GPT-Simplified, each original summary is compared with its GPT-4o-mini output; for Expert-Edited, the same original is compared with the corresponding expert-edited output. BLEU and BERTScore are averaged across aligned pairs, whereas the Flesch-Kincaid scores are computed over the concatenated output texts. SARI is calculated without human references and does not include the standard reference-based \emph{keep} component. Arrows indicate the conventionally preferred direction. All differences are descriptive; no inferential significance tests were conducted.}
\label{tab:simplification-eval}
\end{table}

\paragraph{Source Similarity.}
The GPT outputs receive higher source-similarity scores than the expert-edited outputs, including mean sentence-level BLEU (0.949 vs.\ 0.349) and BERTScore F1 (0.9907 vs.\ 0.9031). These values indicate that the GPT outputs remain substantially closer to the wording of the original summaries. They do not, by themselves, demonstrate higher semantic quality, as deliberate expert rewordings may preserve meaning while reducing lexical overlap.

\paragraph{Simplification Operations.}
The reference-free SARI variant produces a higher score for the GPT simplifications than for the expert-edited summaries (46.76 vs.\ 33.72). Because the calculation uses no human references and does not include the standard reference-based \emph{keep} component, this difference should be interpreted as a descriptive indication of the performed edit operations rather than an overall quality judgment.

\paragraph{Readability.}
The GPT-simplified summaries receive a lower FKGL value (12.10 vs.\ 13.00) and a higher FKRE value (48.54 vs.\ 40.89). The readability formulas therefore favor the GPT outputs in terms of surface-level features. However, these formulas do not assess terminological accuracy, conceptual clarity, or scientific claim calibration.

\paragraph{Qualitative Comparison.}
The expert edits follow a hybrid post-editing strategy. Editors preserve domain-critical terminology, such as \emph{computational lexicon} and \emph{syntactic information}, while selectively adopting clearer LLM-generated formulations for more general explanatory content. This approach aims to improve accessibility without sacrificing scientific precision. An example is shown in Figure~\ref{fig:phase2-response-example}.

\paragraph{Summary.}
Overall, the automatic metrics favor the GPT outputs, partly because they remain closer to the source wording and exhibit simpler surface-level characteristics. The qualitative analysis nevertheless shows how expert post-editing can preserve domain terminology and scientific claim calibration while selectively retaining useful LLM-generated phrasing. As the expert-edited summaries were not independently evaluated by the target readers, these observations should be treated as complementary evidence rather than proof that either condition is uniformly superior.

\begin{figure}[t]
\centering

\begin{minipage}{\linewidth}
\small
\begin{simplificationexample}

{\normalsize\bfseries
COMLEX Syntax: Building a Computational Lexicon
}

\vspace{1em}

{\bfseries (Original)}

\vspace{0.4em}

We describe the design of Comlex Syntax, a
\textbf{computational lexicon} providing detailed
\textbf{syntactic information} for \textbf{approximately}
38,000 English headwords.
\uline{We consider the types of errors which arise in creating such a
lexicon, and how such errors can be measured and controlled.}
Our COMLEX syntax dictionary provides verb subcategorization information
and syntactic paraphrases, but they are indexed by words thus not suitable
to use in generation directly.

\vspace{1em}

{\bfseries (GPT-Simplified)}

\vspace{0.4em}

We explain how we created Comlex Syntax, a
\textbf{computer-based dictionary} that gives detailed
\textbf{grammar information} for \textbf{about} 38,000 main English words.
\uline{We look at the kinds of mistakes that can happen when making this
dictionary, and how we can find and fix these mistakes.}
Our COMLEX syntax dictionary includes details about how verbs are used and
different ways to say things in sentences, but because they are
\textbf{organized by words}, it is not easy to use them for creating new
sentences directly.

\vspace{1em}

{\bfseries (Expert-Edited)}

\vspace{0.4em}

This paper details the design of Comlex Syntax, a
\textbf{computational lexicon} offering extensive
\textbf{syntactic information} for \textbf{approximately}
38,000 English headwords.
\uline{The authors discuss the types of errors that can occur during
lexicon creation and methods for their measurement and control.}
This COMLEX syntax dictionary provides verb subcategorization information
and syntactic paraphrases, but because these are
\textbf{indexed by} individual words, they are not directly suitable for
generating new sentences.

\end{simplificationexample}
\end{minipage}

\caption{Comparison of an original, GPT-simplified, and expert-edited summary from Phase~2.}
\label{fig:phase2-response-example}
\end{figure}

\subsection{Discussion and Implications}

\paragraph{Relation to Prior Work.}
The Phase~1 results are consistent with prior studies showing that LLM-generated simplifications can improve accessibility for readers outside the source domain while introducing risks related to terminology, nuance, and claim strength \cite{engelmann2023textsimplificationscientifictexts,guidroz2025llmbasedtextsimplificationeffect}. The Phase~2 examples further illustrate how expert post-editing can retain or restore domain terminology and calibrated scientific formulations. At the same time, the automatic results reinforce previous criticism of relying on source-overlap metrics alone for text simplification \cite{sulem-etal-2018-bleu}: Higher similarity to the source does not necessarily imply greater accessibility or better preservation of the intended scientific meaning.

\paragraph{Practical Implications.}
The results suggest that LLM-generated simplifications are most useful as initial drafts rather than as directly publishable scientific texts. Feedback from readers outside the source domain can reveal passages that remain difficult, while expert post-editing is needed to verify terminology, preserve meaning, and maintain the appropriate scope and strength of scientific claims.

\paragraph{Research Implications.}
The corpus enables joint evaluation of readability and domain fidelity by linking original summaries with LLM simplifications, reader annotations, and expert-edited references. It can therefore support the development and evaluation of models that simplify scientific texts without compromising terminology or scientific meaning.

\section{Limitations and Challenges}

Although our results demonstrate the potential of LLM-based and human-in-the-loop approaches for scientific text simplification, several limitations remain.

\paragraph{Scale and Coverage.}
Due to participant constraints, expert-edited reference simplifications were created for only 47 of the 1,000 source summaries. This subset may not capture the full linguistic and conceptual diversity of the dataset. Future work should extend the annotations across domains and difficulty levels.

\paragraph{Participant Sampling.}
Phase~1 involved STEM readers without a computer science background, whereas Phase~2 editors were computer science researchers. Although this design reflects the intended distinction between target readers and domain experts, it may introduce sampling bias and does not cover other relevant audiences, such as researchers from the social sciences or humanities, educators, or policy professionals.

\paragraph{Metric Limitations.}
Metrics such as BLEU, BERTScore, SARI, and Flesch-Kincaid capture only selected aspects of similarity and readability. They do not adequately measure semantic nuance, rhetorical clarity, or factual fidelity, which are central to scientific text simplification. Moreover, our analysis reports aggregate values without paired significance tests, and the SARI evaluation is reference-free. The expert-edited summaries were also not evaluated independently by target readers. The reported differences should therefore be interpreted descriptively.

\paragraph{Domain and Language Scope.}
The study is limited to English-language computer science texts. As scientific terminology and communication practices differ across disciplines and languages, the results may not generalize directly to other settings. Extending the workflow to multilingual corpora and additional scientific domains would improve its broader applicability.

\paragraph{Model Dependence.}
The initial simplifications were generated using GPT-4o-mini and a single fixed zero-shot prompt. Different models, prompts, or decoding strategies may produce different results. Systematic comparisons across models and prompting approaches remain future work.

\paragraph{Annotation Reliability.}
Phase~1 annotations reflect perceived difficulty and may therefore contain noise. In particular, participants may conflate unfamiliar terminology with linguistic complexity. Expert post-editing partly mitigates this limitation, but the reliability of the annotations should be examined more systematically.

\paragraph{Ethical Considerations.}
Text simplification may unintentionally distort scientific claims, for example by overstating findings or omitting methodological qualifications. Expert validation reduces this risk but cannot eliminate it. Simplified texts should therefore be used with particular care in educational, public-facing, and policy-related contexts.

\section{Conclusion}

This work introduced a human-in-the-loop framework for simplifying scientific summaries with large language models (LLMs), addressing accessibility barriers in interdisciplinary research communication. Starting from expert-written summaries in SciSummNet, we generated baseline simplifications using GPT-4o-mini and engaged cross-disciplinary STEM readers to identify linguistic and conceptual difficulties. Their annotations informed a second phase in which domain experts revised the LLM outputs to produce expert-edited reference simplifications.

Phase~1 shows that readers generally preferred the GPT-generated versions for understanding and simplicity, while also identifying cases of awkward phrasing and reduced technical precision. In Phase~2, the automatic metrics generally favor the GPT outputs because they remain closer to the source, whereas qualitative inspection illustrates how expert post-editing can preserve terminology and scientific claim calibration. These findings motivate treating LLM simplifications as drafts that benefit from feedback by cross-disciplinary readers and domain-informed review.

The resulting corpus includes original summaries, LLM-generated simplifications, reader annotations, and expert-edited references, providing a reproducible resource and evaluation setting for scientific text simplification.

Future work will extend the number and disciplinary coverage of the expert-edited summaries, evaluate these summaries with independent target readers, and compare alternative models and prompting strategies. Further work should also develop evaluation methods that jointly capture accessibility, factual fidelity, terminology preservation, and scientific claim calibration.

\section*{Data and Code Availability}
The corpus, annotation artifacts, and code are publicly available at: 
\url{https://github.com/faerber-lab/scientific-text-simplification-corpus}

\section*{Ethics Statement}

The study involved cross-disciplinary STEM readers and domain experts. All participants provided informed consent and could withdraw at any time. We collected no personally identifying information. As text simplification may alter scientific claims, our guidelines and expert review explicitly focused on preserving meaning and terminology.

\section*{Acknowledgements}

This research was funded by the Center for Scalable Data Analytics and Artificial Intelligence (ScaDS.AI) and the German Federal Ministry of Research, Technology and Space (BMFTR) through the Software Campus project (01IS23070). We thank the reviewers and Shuzhou Yuan for valuable feedback.

\bibliography{references} %

\end{document}